\PassOptionsToPackage{unicode}{hyperref}
\PassOptionsToPackage{hyphens}{url}
\PassOptionsToPackage{dvipsnames,svgnames,x11names}{xcolor}
\documentclass[
]{article}
\usepackage{xcolor}
\usepackage{amsmath,amssymb}
\usepackage{iftex}
\ifPDFTeX
  \usepackage[T1]{fontenc}
  \usepackage[utf8]{inputenc}
  \usepackage{textcomp} % provide euro and other symbols
\else % if luatex or xetex
  \usepackage{unicode-math} % this also loads fontspec
  \defaultfontfeatures{Scale=MatchLowercase}
  \defaultfontfeatures[\rmfamily]{Ligatures=TeX,Scale=1}
\fi
\usepackage{lmodern}
\ifPDFTeX\else
\fi
\IfFileExists{upquote.sty}{\usepackage{upquote}}{}
\IfFileExists{microtype.sty}{% use microtype if available
  \usepackage[]{microtype}
  \UseMicrotypeSet[protrusion]{basicmath} % disable protrusion for tt fonts
}{}
\makeatletter
\@ifundefined{KOMAClassName}{% if non-KOMA class
  \IfFileExists{parskip.sty}{%
    \usepackage{parskip}
  }{% else
    \setlength{\parindent}{0pt}
    \setlength{\parskip}{6pt plus 2pt minus 1pt}}
}{% if KOMA class
  \KOMAoptions{parskip=half}}
\makeatother
\makeatletter
\ifx\paragraph\undefined\else
  \let\oldparagraph\paragraph
  \renewcommand{\paragraph}{
    \@ifstar
      \xxxParagraphStar
      \xxxParagraphNoStar
  }
  \newcommand{\xxxParagraphStar}[1]{\oldparagraph*{#1}\mbox{}}
  \newcommand{\xxxParagraphNoStar}[1]{\oldparagraph{#1}\mbox{}}
\fi
\ifx\subparagraph\undefined\else
  \let\oldsubparagraph\subparagraph
  \renewcommand{\subparagraph}{
    \@ifstar
      \xxxSubParagraphStar
      \xxxSubParagraphNoStar
  }
  \newcommand{\xxxSubParagraphStar}[1]{\oldsubparagraph*{#1}\mbox{}}
  \newcommand{\xxxSubParagraphNoStar}[1]{\oldsubparagraph{#1}\mbox{}}
\fi
\makeatother

\usepackage{longtable,booktabs,array}
\usepackage{calc} % for calculating minipage widths
\usepackage{etoolbox}
\makeatletter
\patchcmd\longtable{\par}{\if@noskipsec\mbox{}\fi\par}{}{}
\makeatother
\IfFileExists{footnotehyper.sty}{\usepackage{footnotehyper}}{\usepackage{footnote}}
\makesavenoteenv{longtable}
\usepackage{graphicx}
\makeatletter
\newsavebox\pandoc@box
\newcommand*\pandocbounded[1]{% scales image to fit in text height/width
  \sbox\pandoc@box{#1}%
  \Gscale@div\@tempa{\textheight}{\dimexpr\ht\pandoc@box+\dp\pandoc@box\relax}%
  \Gscale@div\@tempb{\linewidth}{\wd\pandoc@box}%
  \ifdim\@tempb\p@<\@tempa\p@\let\@tempa\@tempb\fi% select the smaller of both
  \ifdim\@tempa\p@<\p@\scalebox{\@tempa}{\usebox\pandoc@box}%
  \else\usebox{\pandoc@box}%
  \fi%
}
\def\fps@figure{htbp}
\makeatother

\NewDocumentCommand\citeproctext{}{}
\NewDocumentCommand\citeproc{mm}{%
  \begingroup\def\citeproctext{#2}\cite{#1}\endgroup}
\makeatletter
 \let\@cite@ofmt\@firstofone
 \def\@biblabel#1{}
 \def\@cite#1#2{{#1\if@tempswa , #2\fi}}
\makeatother
\newlength{\cslhangindent}
\newlength{\csllabelwidth}
\newenvironment{CSLReferences}[2] % #1 hanging-indent, #2 entry-spacing
 {\begin{list}{}{%
  \setlength{\itemindent}{0pt}
  \setlength{\leftmargin}{0pt}
  \setlength{\parsep}{0pt}
  \ifodd #1
   \setlength{\leftmargin}{\cslhangindent}
   \setlength{\itemindent}{-1\cslhangindent}
  \fi
  \setlength{\itemsep}{#2\baselineskip}}}
 {\end{list}}
\usepackage{calc}

\usepackage{arxiv}
\usepackage{orcidlink}
\usepackage{amsmath}
\usepackage{multirow}
\usepackage{booktabs}
\usepackage{arydshln}
\usepackage{siunitx}
\usepackage[T1]{fontenc}
\usepackage{arxiv}
\usepackage{orcidlink}
\usepackage{amsmath}
\usepackage{multirow}
\usepackage{booktabs}
\usepackage{arydshln}
\usepackage{algorithm}
\usepackage{algpseudocode}
\usepackage[T1]{fontenc}
\makeatletter
\@ifpackageloaded{caption}{}{\usepackage{caption}}
\AtBeginDocument{%
\ifdefined\contentsname
  \renewcommand*\contentsname{Table of contents}
\else
  \newcommand\contentsname{Table of contents}
\fi
\ifdefined\listfigurename
  \renewcommand*\listfigurename{List of Figures}
\else
  \newcommand\listfigurename{List of Figures}
\fi
\ifdefined\listtablename
  \renewcommand*\listtablename{List of Tables}
\else
  \newcommand\listtablename{List of Tables}
\fi
\ifdefined\figurename
  \renewcommand*\figurename{Figure}
\else
  \newcommand\figurename{Figure}
\fi
\ifdefined\tablename
  \renewcommand*\tablename{Table}
\else
  \newcommand\tablename{Table}
\fi
}
\@ifpackageloaded{float}{}{\usepackage{float}}
\floatstyle{ruled}
\@ifundefined{c@chapter}{\newfloat{codelisting}{h}{lop}}{\newfloat{codelisting}{h}{lop}[chapter]}
\floatname{codelisting}{Listing}

\makeatother
\makeatletter
\@ifpackageloaded{caption}{}{\usepackage{caption}}
\@ifpackageloaded{subcaption}{}{\usepackage{subcaption}}
\makeatother
\makeatletter
\@ifpackageloaded{algorithm}{}{\usepackage{algorithm}}
\makeatother
\makeatletter
\@ifpackageloaded{algpseudocode}{}{\usepackage{algpseudocode}}
\makeatother
\makeatletter
\@ifpackageloaded{caption}{}{\usepackage{caption}}
\makeatother
\usepackage{bookmark}
\IfFileExists{xurl.sty}{\usepackage{xurl}}{} % add URL line breaks if available
\hypersetup{
  pdftitle={L0: Reinforcement Learning to Become General Agents},
  pdfkeywords={AI Agents, Reinforcement
Learning, Code-as-Action, Verifiable Rewards},
  colorlinks=true,
  linkcolor={blue},
  filecolor={Maroon},
  citecolor={Blue},
  urlcolor={Blue},
  pdfcreator={LaTeX via pandoc}}

\title{L0: Reinforcement Learning to Become General Agents}
\author{
  \textbf{Junjie Zhang\thanks{Equal Contribution}\ }, \quad
  \textbf{Jingyi Xi\footnotemark[1]\ }, \quad
  \textbf{Zhuoyang Song\footnotemark[1]\ }, \quad
\\
  \textbf{Junyu Lu}, \quad
  \textbf{Yuhua Ke},  \quad
  \textbf{Ting Sun}, \quad
  \textbf{Yukun Yang}, \quad
\\
  \textbf{Jiaxing Zhang\thanks{Corresponding Author}\ }, \quad
  \textbf{Songxin Zhang\footnotemark[2]\ }, \quad
  \textbf{Zejian Xie\footnotemark[2]}
\\
\\
  Lionrock AI Lab, China Merchants Research Institute of Advanced Technology
}

\begin{document}
\maketitle
\begin{abstract}
Training large language models (LLMs) to act as autonomous agents for
multi-turn, long-horizon tasks remains significant challenges in
scalability and training efficiency. To address this, we introduce
\textbf{L-Zero (L0)}, a scalable, end-to-end training pipeline for
general-purpose agents. Featuring a low-cost, extensible, and sandboxed
concurrent agent worker pool, L0 lowers the barrier for applying
reinforcement learning in complex environments. We also introduce
NB-Agent, the agent scaffold within L0, which operates in a
``code-as-action'' fashion via a Read-Eval-Print-Loop (REPL). We
evaluate L0 on factuality question-answering benchmarks. Our experiments
demonstrate that a base model can develop robust problem-solving skills
using solely Reinforcement Learning with Verifiable Rewards (RLVR). On
the Qwen2.5-7B-Instruct model, our method boosts accuracy on
\emph{SimpleQA} from 30 \% to 80 \% and on \emph{HotpotQA} from 22 \% to
41 \%. We have open-sourced the entire L0 system, including our L0
series models, the NB-Agent, a complete training pipeline, and the
corresponding training recipes on
\href{https://github.com/cmriat/l0}{GitHub}.
\end{abstract}
{\bfseries \emph Keywords}
\def\sep{\textbullet\ }
AI Agents \sep Reinforcement Learning \sep Code-as-Action \sep 
Verifiable Rewards

\floatname{algorithm}{Algorithm}

\section{Introduction}\label{introduction}

Recent progress in reinforcement learning
(\citeproc{ref-wang2025ragen}{Z. Wang et al. 2025};
\citeproc{ref-feng2025gigpo}{Feng et al. 2025}) has enabled Large
Language Models (LLMs) to become active agents, capable of performing
complex, multi-step tasks by interacting with external environments and
tools. Pioneering works have demonstrated impressive capabilities by
integrating search functionalities with the models' intrinsic reasoning,
allowing agents to dynamically query for information and ground their
responses in external knowledge. For instance, Search-o1
(\citeproc{ref-li2025search-o1}{Li et al. 2025}) incorporates an agentic
retrieval-augmented generation (RAG) mechanism so that the model
dynamically retrieves external knowledge whenever it encounters
uncertainty. Search‐R1 (\citeproc{ref-jin2025search-r1}{Jin et al.
2025}) trains an LLM to generate multiple search queries during
step-by-step reasoning and retrieve real-time information, yielding
large accuracy gains on QA tasks.

However, existing works have severe limitations in multi-turn RL
pipelines, as they frame long-horizon reasoning as a single-step bandit
problem with only final-answer rewards, ignoring intermediate signals.
Similarly, strict prompt templates are often used that allow just one
tool call per reasoning turn. Such designs make it difficult for the
agent to coordinate multiple tools or refine its behavior through
turn-level feedback. In practice, real-world agentic tasks typically
require orchestrating heterogeneous actions (e.g.~web search, database
queries, code execution) over many steps and preserving context or
memory across those steps. It indicates that training these agents for
long-horizon, stateful interactions remains a significant challenge,
demanding both expressive agent architectures capable of managing
internal state and scalable RL infrastructure that can handle
resource-intensive, multi-turn rollouts. This prevailing gap motivates
our research toward bridging advanced agent designs and effective,
scalable reinforcement learning frameworks.

To address this prevailing gap, we introduce L-Zero (L0), a scalable,
end-to-end pipeline for training general-purpose agents. We first
present the architectural heart of our system, the NB-Agent, a novel
agent scaffold that operates using a ``code-as-action'' paradigm
(\citeproc{ref-wang2024codeact}{X. Wang et al. 2024}) within an
interactive, stateful Python environment. We then detail the end-to-end
reinforcement learning framework designed to train this agent,
highlighting three core innovations: an agentic policy gradient tailored
for complex, multi-token actions; a multi-faceted reward model based on
verifiable outcomes; and a highly scalable, sandboxed infrastructure
engineered for robust, parallelized agent rollouts. Finally, we present
a comprehensive empirical evaluation of L0 on several challenging
question-answering benchmarks. Our experiments demonstrate that the
agentic scaffold alone provides a strong structural prior for reasoning,
which is then significantly amplified by our RL training, leading to
substantial performance gains over strong supervised baselines.

\section{NB-Agent}\label{nb-agent}

\begin{figure}

\centering{

\pandocbounded{\includegraphics[keepaspectratio]{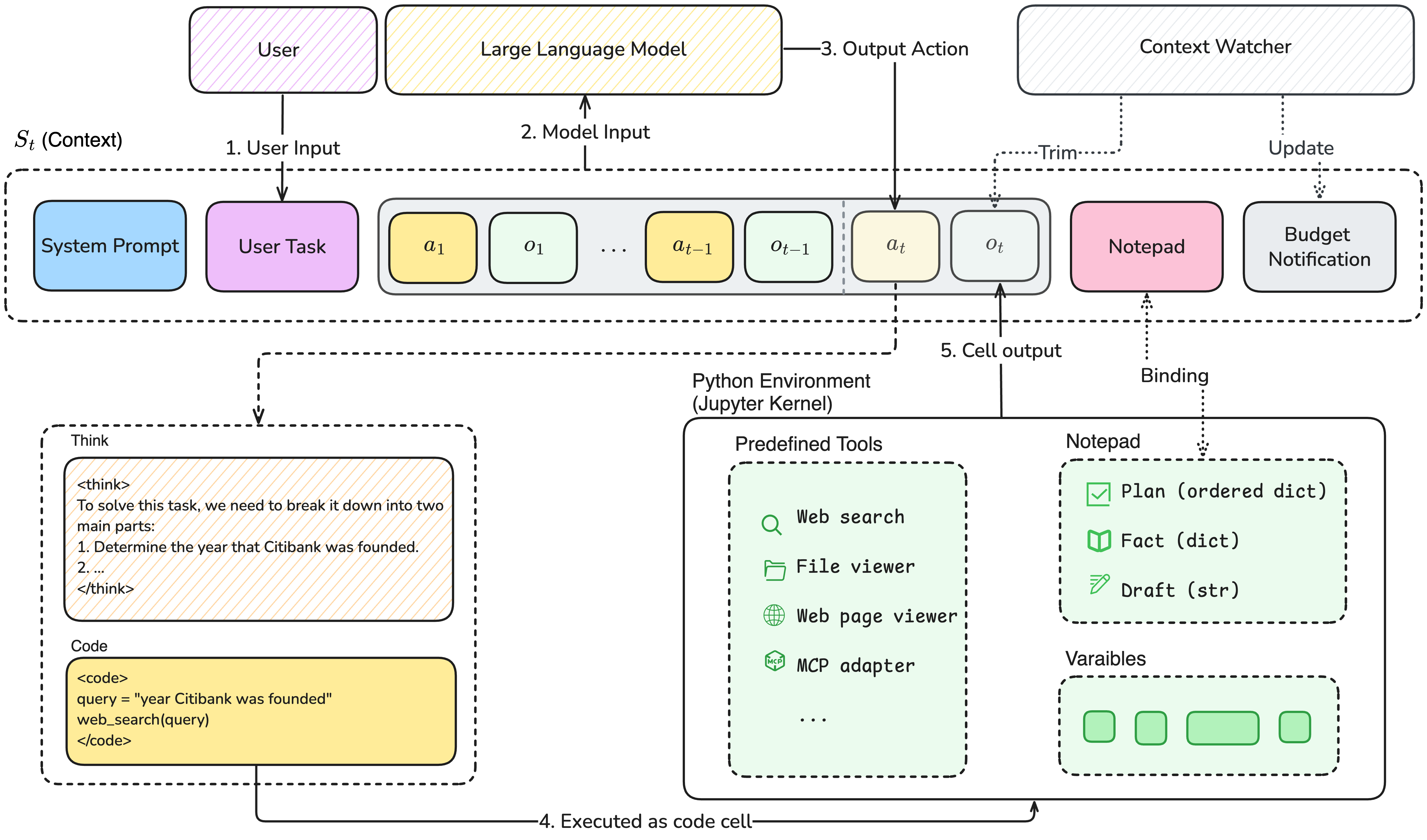}}

}

\caption{\label{fig-main}The overall architecture of NB Agent, showing a
REPL-style Think-Code-Observe loop. The LLM generates a reasoning trace
(\texttt{\textless{}think\textgreater{}}) and Python code
(\texttt{\textless{}code\textgreater{}}), executed in a Jupyter kernel.
The execution output (\texttt{\textless{}output\textgreater{}}) becomes
the next observation. The cycle repeats until a final answer.}

\end{figure}%

Inspired by prior work that champions code as a general-purpose action
space for autonomous agents, we introduce the NB-Agent. Our agent
operates within a cyclical process as detailed in Algorithm
\ref{alg:agentic_loop} and Figure~\ref{fig-main}. The loop initiates
with a user-defined task and continues until a definitive answer is
submitted.

\begin{algorithm}
\caption{Agentic Loop}
\label{alg:agentic_loop}
\begin{algorithmic}
\State {\bfseries Input:} Initial state $s_0 = [\text{system\_prompt, user\_task}]$
\State {\bfseries Ensure:} Final result when task is completed
\State $t \gets 1$
\While{True}
    \State $s_t \gets \text{context\_watcher}(s_{t-1}, a_{t-1}, o_{t-1})$
    \State $a_t \gets (a_{\text{think},t}, a_{\text{code},t}) \gets \Pi_{\theta}(s_t)$
    \State $o_t \gets \text{execute\_code\_in\_environment}(s_t, a_t)$
    \If{is\_final\_answer\_submitted($o_t$)}
        \State {\bfseries break}
    \EndIf
    \State $t \gets t + 1$
\EndWhile
\State {\bfseries return} final result
\end{algorithmic}
\end{algorithm}

As illustrated, upon receiving a task, the NB-Agent enters a
``Think-Code-Observe'' loop. In each cycle, the LLM generates a
\texttt{\textless{}think\textgreater{}} trace to reason about the next
steps, followed by a \texttt{\textless{}code\textgreater{}} block
containing Python code. This code is then executed within a code cell,
mimicking a human's interaction with a Jupyter Notebook. The output from
the cell execution is captured, wrapped in an
\texttt{\textless{}output\textgreater{}} tag, and serves as the
observation for the next cycle.

The \texttt{context\_watcher} module is responsible for managing the
context window of the LLM. If the conversational history exceeds the
context length, it will strategically truncate previous actions and
outputs. A notification regarding the remaining token and step budget is
also sent to the LLM to inform its planning. This loop continues until
the agent invokes the \texttt{submit\_final\_answer} tool to terminate
the process and return the result.

\subsection{Agent's Environment: An Interactive Python
Kernel}\label{agents-environment-an-interactive-python-kernel}

We ground our agent's operation within a standard Python environment,
specifically a Jupyter kernel. This design choice is pivotal for two
reasons. First, it provides a robust and stateful execution environment
where the agent can define variables, install new libraries, and
maintain a persistent session state, mirroring the workflow of a human
data scientist.

Second, and more critically, it leverages the Read-Eval-Print Loop
(REPL) as a natural and effective mechanism for the agent's interaction
cycle. The agent's generated code (read) is executed by the kernel
(eval), and the output is captured (print) and returned as an
observation. This ``action-observe'' paradigm inherent to REPL provides
immediate feedback, allowing the agent to iteratively refine its
approach based on concrete results.

\subsection{Overcoming Context
Limitations}\label{overcoming-context-limitations}

A primary challenge for LLM agents is the finite context window. We
introduce two key innovations to address this:

\textbf{Context-Variable Binding via a ``Notepad''}: We establish a
bidirectional link between the agent's transient context and persistent
Python variables through a structured Notepad tool. This Notepad is a
mutable object within the Python environment, comprising dedicated
modules for planning, fact storage, and drafting. By manipulating this
object through code, the agent can actively control what information is
preserved beyond the immediate context window, effectively giving it a
manageable, long-term memory.

\textbf{REPL-driven Information Retrieval}: The REPL interface allows
the agent to not only execute actions but also to manage its memory. The
agent can save critical information (e.g., API results, intermediate
findings) to Python variables using simple commands and can retrieve it
in later steps as needed. This mechanism transforms the Python
environment into a dynamic and accessible external memory, enabling the
agent to handle long-term dependencies and complex data structures
efficiently.

\section{End-to-End Agentic Reinforcement
Learning}\label{end-to-end-agentic-reinforcement-learning}

\begin{figure}

\centering{

\pandocbounded{\includegraphics[keepaspectratio]{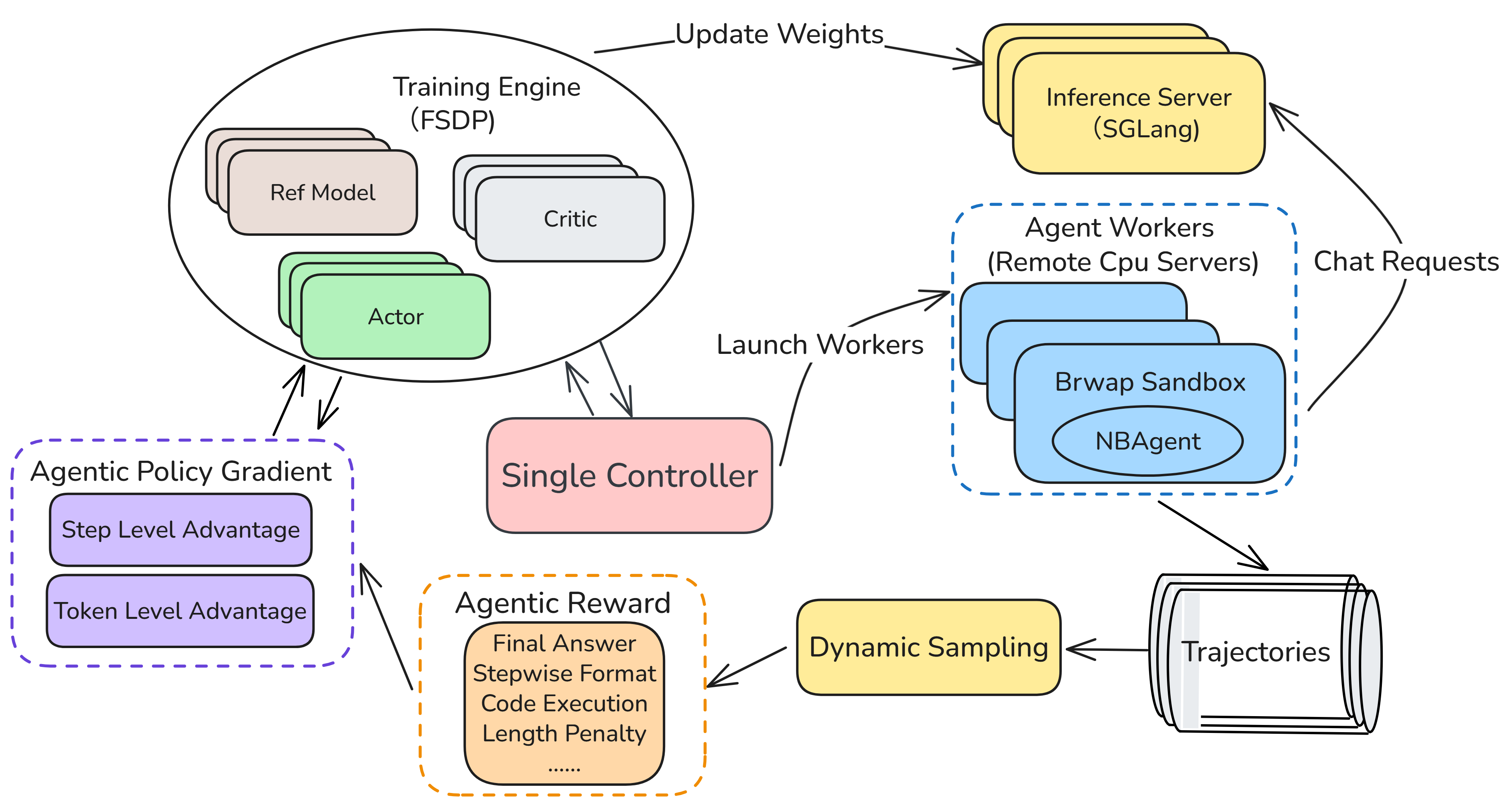}}

}

\caption{\label{fig-agentrl}The AgentRL training architecture. A central
controller dispatches tasks to many isolated agent workers and a shared
inference server hosting the NB-Agent. The training engine collects
trajectories and updates the NB-Agent via agentic policy gradient using
verifiable rewards.}

\end{figure}

Training the NB-Agent with RLVR requires a framework built specifically
for long-horizon, interactive tasks. To this end, we develop a
comprehensive framework for end-to-end agentic reinforcement learning
based on verl. As illustrated in Figure~\ref{fig-agentrl}, our approach
is built on three core innovations designed to overcome the unique
challenges of this domain:

\begin{itemize}
\item
  A novel agentic policy gradient that redefines an ``action'' to
  encompass a entire sequence of reasoning and code generation.
\item
  A verifiable reward model that provides multi-faceted, automated
  feedback on final answer correctness, format compliance, and code
  execution.
\item
  A scalable, decoupled infrastructure engineered to handle the
  resource-intensive demands of parallel, multi-turn agent rollouts.
\end{itemize}

\subsection{Agentic Policy Gradient}\label{agentic-policy-gradient}

In contrast to conventional language modeling where an action
corresponds to generating a single token, our agentic framework defines
an action as a complete sequence of tokens that constitutes a meaningful
step. This necessitates a corresponding adaptation of the policy
gradient method. Therefore, we formulate a policy gradient that operates
at the level of these action sequences. The policy gradient,
\(\hat{g}\), is estimated as the expectation over timesteps \(t\):
\[\hat{g} = \hat{\mathbb{E}}_t[\nabla \log \pi_{\theta}(a_t, s_t) \hat{A}_t].\]

Since \(\Pi_{\theta}(a_t, s_t)\) is the probability of generating an
action sequence under state \(s_t\), this can be decomposed into the
product of conditional probabilities for each token in the sequence:

\[\pi_{\theta}(a_t, s_t) = P_{\theta}(a_t | s_t) = \prod_{i=1}^{|a_t|} P_{\theta}(tok_i| [s_{t}, tok_0, ..., tok_{i-1}]),\]

where \(tok_i\) is the \(i\)-th token generated in \(a_t\). Then, the
gradient can be rewritten as:
\[\hat{g} = \hat{\mathbb{E}}_t[\nabla \log \prod_{i=1}^{|a_t|} P_{\theta}(tok_i| [s_{t}, tok_0, ..., tok_{i-1}]) \hat{A}_t]\]
\[ = \hat{\mathbb{E}}_t[\sum_{i=1}^{|a_t|} \nabla \log P_{\theta}(tok_i| [s_{t}, tok_0, ..., tok_{i-1}]) \hat{A}_t].\]

In practice, we adopt the token-level normalization strategy proposed by
DAPO (\citeproc{ref-yu2025dapo}{Yu et al. 2025}) to address the
contribution of longer trajectories:

\[\hat{g} = \frac{1}{\sum_{j=1}^{N}\sum_{t=1}^{T_n}|a_t|} \sum_{n=1}^{N}\sum_{t=1}^{T_n} \hat{A}_t \sum_{i=1}^{|a_t|} \nabla \log P_{\theta}(tok_i| [s_{t}, tok_0, ..., tok_{i-1}]),\]

where \(N\) is the number of samples in a batch, \(T_n\) is number of
steps of \(n\)-th trajectory, and \(|a_t|\) is the length of the action
sequence at time step \(t\). For the advantage estimation,
\(\hat{A}_t\), we adopt the approach from REINFORCE++
(\citeproc{ref-hu2025reinforce}{Hu 2025}):

\[\hat{A}_t = \sum_{t'=t}^{T} \gamma^{t'-t} r_{t'},\]

where \(r_{t'}\) is the reward at time step \(t'\) and \(\gamma\) is the
discount factor, and we do a batch-step-wise advantage normalization to
stabilize the training:
\[\hat{A}_t = \frac{\hat{A}_t - \mathbb{E}[\hat{A}_t]}{\sqrt{\text{Var}[\hat{A}_t] + \epsilon}},\]

where the expectation and variance are computed over all timesteps of
all trajectories within the batch, and \(\epsilon\) is a small constant
to avoid division by zero.

Our implementation adheres to a strictly on-policy training regime,
meaning we do not use corrective mechanisms like importance sampling or
PPO-style clipping. Furthermore, in line with common practices, we
include an explicit KL-divergence penalty term in the objective
function.

To enhance exploration and stabilize training, we employ a dynamic
sampling strategy inspired by DAPO. This procedure involves generating
multiple trajectories from the same input query and randomly discard
trajectories that yield either zero or maximum core rewards.

\subsection{Verified Reward for Agentic
RL}\label{verified-reward-for-agentic-rl}

To objectively measure the quality of a trajectory
\(\tau = {(s_1,a_1), \dots, (s_T,a_T)}\), we design a verifiable reward
\(R(\tau)\) that combines multiple criteria for multi-step agentic
training. This reward is verifiable in the sense that it is computed
automatically from the agent's outputs and environment feedback, without
requiring human judgments. Formally, \(R(\tau)\) is composed of three
components:

\textbf{Final Answer Correctness}: This term rewards the agent for the
quality of its submitted answer. For tasks with a ground-truth solution,
the reward, \(R_{final}\), is calculated as a weighted sum of three
metrics:

\[R_{final} = 0.9 \cdot \mathbf{1}_{\text{exact match}} + 0.1 \cdot \mathbf{1}_{\text{has answer}}\]

where \textbf{\(\mathbf{1}_{\text{exact match}}\)}(EM score) is a binary
indicator that is 1 if the agent submits an exact match answer and 0
otherwise. \textbf{\(\mathbf{1}_{\text{has answer}}\)} is a binary
indicator that is 1 if the agent submits an answer and 0 otherwise.

This composite score provides a nuanced signal that strongly rewards
perfect answers (EM weight), and lightly encourages the agent to provide
a response rather than terminating without one.

\textbf{Stepwise Format Compliance:} This term gives reward for adhering
to the required
\texttt{\textless{}think\textgreater{}\textless{}/think\textgreater{}}and
\texttt{\textless{}code\textgreater{}\textless{}/code\textgreater{}}
output structure at each step. The agent is incentivized to output
well-formed reasoning and code blocks that match the template
(e.g.~properly using the tags and not mixing thought text into code).

\textbf{Code Execution \& Correctness:} This component rewards the agent
for producing code actions that execute successfully and achieve their
intended effect. If a \texttt{\textless{}code\textgreater{}} action
compiles/runs without errors, the reward is positive.

\subsection{Scalable Agent RL Infra}\label{scalable-agent-rl-infra}

Training agents for long-horizon tasks that involve real-world
interactions and code execution introduces significant infrastructure
challenges, particularly concerning resource allocation and variable
task completion times. To address this, we developed a scalable and
robust infrastructure engineered with several key design principles:

\begin{itemize}
\item
  \textbf{Decoupled Architecture:} Our infrastructure is built on a
  decoupled architecture where \textbf{a pool of Agent Workers}, managed
  on remote CPU servers, handles all environment interaction and
  trajectory collection. This pool's workload is precisely controlled
  via a \texttt{max\_concurrency} parameter to manage system load.
  Workers communicate with a dedicated Inference Server, running SGLang,
  via stateless HTTP requests to query the latest policy for actions.
  This separation of concerns---CPU-based environment simulation from
  GPU-based inference and training---prevents computational bottlenecks
  and allows each component to be scaled independently.
\item
  \textbf{Lightweight, Isolated Agent Environments:} Each Agent Worker
  is encapsulated within \textbf{a secure sandbox using Bubblewrap}.
  This technology provides strong process isolation with minimal
  performance overhead. Critically, unlike traditional container
  solutions like Docker, Bubblewrap operates without requiring root
  privileges, which both enhances security and simplifies deployment
  across the distributed worker pool, making it ideal for large-scale
  parallelization.
\item
  \textbf{Agent Execution Control:} To ensure system stability and
  consistent data throughput despite variations in task complexity, we
  implement \textbf{robust execution controls}. These include a timeout
  mechanism that terminates workers exceeding a \texttt{time\ threshold}
  and a \texttt{maximum\ retry} limitation for tasks that repeatedly
  fail.
\end{itemize}

Collectively, these design choices create a high-throughput, scalable,
and fault-tolerant agent rollout pipeline, providing the stable
foundation required for effective large-scale reinforcement learning.

\section{Experiment}\label{experiment}

\subsection{Evaluation Datasets and
Metrics}\label{evaluation-datasets-and-metrics}

We comprehensively evaluate L0 on a set of open-domain multi-hop
question answering tasks, categorized as follows: \textbf{HotpotQA}
(\citeproc{ref-yang2018hotpotqa}{Z. Yang et al. 2018}) represents the
first large-scale dataset designed explicitly for multi-hop reasoning,
requiring models to reason across multiple Wikipedia paragraphs;
\textbf{2WikiMultiHopQA} (\citeproc{ref-ho20202wiki}{Ho et al. 2020})
explicitly provides reasoning paths for assessing multi-step inference
capabilities; \textbf{Musique}
(\citeproc{ref-trivedi2022musique}{Trivedi et al. 2022}) consists of
2-to-4-hop reasoning questions constructed from five existing single-hop
datasets; \textbf{Bamboogle} (\citeproc{ref-press2023bamboogle}{Press et
al. 2023}) comprises complex queries that Google frequently answers
incorrectly, intended to test models' cross-domain combinational
reasoning abilities; \textbf{SimpleQA}
(\citeproc{ref-wei2024simpleqa}{Wei et al. 2024}) focus on short,
fact-seeking queries.

We adopt Exact Match (EM), F1-score and LLM-as-a-judge (LJ) as our
evaluation metric, considering a prediction correct if its normalized
form exactly matches any one of the normalized ground-truth answers.

\subsection{Baselines}\label{baselines}

To systematically evaluate the effectiveness of L0, we compare our
approach against three categories of baseline methods:

\textbf{Basic Prompting Engineering:} This category includes Direct
Prompting, Chain-of-Thought (CoT), and standard Retrieval-Augmented
Generation (RAG).

\textbf{Advanced RAG Methods:} We consider RAgent
(\citeproc{ref-jayasundara2024ragent}{Jayasundara, Arachchilage, and
Russello 2024}) and Search-o1 (\citeproc{ref-li2025search-o1}{Li et al.
2025}). RAgent employs a retrieval-based agentic policy-generation
framework to systematically organize internal knowledge, while Search-o1
integrates an agent-based search workflow into large-scale reasoning
processes.

\textbf{Reinforcement Learning-based Methods:} This category includes
DeepSeek-R1 (\citeproc{ref-guo2025deepseek-r1}{Guo et al. 2025}),
Search-R1 (\citeproc{ref-jin2025search-r1}{Jin et al. 2025}), and
ZeroSearch (\citeproc{ref-sun2025zerosearch}{Sun et al. 2025}). In
DeepSeek-R1, the policy model performs deep reasoning relying solely on
internal knowledge. Search-R1 allows the policy model to interact
repeatedly with an actual search engine during inference. ZeroSearch
further incorporates simulated search during training to incentivize the
LLM's ability to leverage real-world search engines effectively.

During the evaluation process, all baseline methods consistently employ
SerpAPI as the search engine, with their performance and experoment
settings adopted from those reported in ZeroSearch.

\subsection{Experiment Settings}\label{experiment-settings}

We employ Qwen2.5-7B-Instruct, Qwen2.5-32B-Instruct and
Qwen3-4B-Thinking (\citeproc{ref-yang2025qwen3}{A. Yang et al. 2025}) as
backbone models, configuring the generation parameters with a max-tokens
of 32,768 and a top-p sampling threshold of 0.9. To simulate realistic
retrieval scenarios, we use Google Web Search via SerpAPI as the
external search engine and retrieve webpage content using the Jina
Reader API. Furthermore, we introduce a novel data filtering and
stratification strategy based on objectivity, temporal stability, and
question difficulty, selecting a total of 20K QA pairs from the training
sets of 2WikiMultihopQA (\citeproc{ref-ho20202wiki}{Ho et al. 2020}),
TriviaQA (\citeproc{ref-joshi2017triviaqa}{Joshi et al. 2017}), NQ
(\citeproc{ref-kwiatkowski2019nq}{Kwiatkowski et al. 2019}), and
HotpotQA (\citeproc{ref-yang2018hotpotqa}{Z. Yang et al. 2018}) datasets
for reinforcement learning.

\subsection{Main Result}\label{main-result}

In Table Table~\ref{tbl-qa-results-table1}, we report two configurations
of our method: L0-Scaffold (NB-Agent + Qwen, without RL training), and
L0-RL (NB-Agent + Qwen, training with AgentRL).

The L0-Scaffold configuration instantiates the NB-Agent's structured
reasoning template but without RL training, beating the Direct Answer
(11.87\%) and RAG (17.33\%) on average. This scaffold explicitly
separates reasoning, query generation, and answering in a multi-turn
loop, which enforces a structured decision-making process to enhance the
transparency and reliability. It is noteworthy that due to the inherent
complexity of the Scaffold architecture, the performance prior to RL
training is comparable to that of Search-o1.

L0-RL further applies reinforcement learning to boost performance by
teaching the model to use the equipped tools and memory management
effectively. Compared with L0-Scaffold, L0-RL demonstrates a substantial
performance improvement, with the average score dramatically increasing
from 18.57\% to 38.28\%. By training the agent to learn from the
verifiable outcomes of its self-generated code, AgentRL teaches it to
master its own tools, manage memory effectively, and develop
sophisticated, self-correcting strategies. The agent evolves from a
structured but naive operator into a highly autonomous and proficient
problem-solver.

L0-RL's gains compare favorably to other RL-driven LLM Agents,
outperforming the Search-R1 and ZeroSearch by a clear margin on average,
indicating that the L0 scaffold provides a far richer and more
expressive environment for reinforcement learning. While other methods
train agents to learn when to call a single tool (e.g., a search
engine), our framework trains the agent to become a programmatic
problem-solver, learning how to compose actions, manage state, and
reason within a structured environment. This synergy between an advanced
agent architecture and a tailored RL training process enables a more
capable and generalizable form of agentic intelligence.

Finally, we analyze performance across three different backbone models
to understand the effects of scale and inherent model capabilities, with
results shown in Table Table~\ref{tbl-qa-results-table2}. As expected,
scaling the model size from 7B to 32B elevates the baseline performance
of L0-Scaffold from 20.88\% to 43.50\% on EM score, showing that a more
stronger foundation model is better able to leverage the structured
environment provided by the scaffold. More interestingly, the
Qwen-3-4B-Thinking model, which is pre-trained for robust tool-use
capabilities, shows the most pronounced improvement after RL training.
Although its L0-Scaffold score starts lower (14.78\% on EM and 18.65\%
on F1), it achieves an L0-RL score of 44.67\% on EM and 54.36\% on F1.
This suggests a powerful synergy that models with inherent reasoning and
tool-use abilities are exceptionally receptive to AgentRL training. They
can more effectively utilize the fine-grained learning signals to unlock
their full potential, resulting in even more dramatic performance
enhancements.

\begin{table}

\caption{\label{tbl-qa-results-table1}Main results of different methods. The best performance is set in bold.}

\centering{

\centering

\begin{tabular}{l S[table-format=2.2]
                   S[table-format=2.2]
                   S[table-format=2.2]
                   S[table-format=2.2]}
\toprule
\multirow{2}{*}{\textbf{Method}} & \multicolumn{4}{c}{\textbf{Multi-Hop QA}} \\
\cmidrule(l){2-5}
 & {\textbf{HotpotQA}} & {\textbf{Musique}} & {\textbf{Bamboogle}} & {\textbf{Avg.}} \\
\midrule
\multicolumn{5}{l}{\textbf{\textit{Qwen-2.5-7B-Base/Instruct}}} \\
\addlinespace
Direct Answer      & 16.40 & 4.80  & 14.40 & 11.87 \\
CoT                & 16.20 & 6.60  & 24.00 & 15.60 \\
RAG                & 25.80 & 9.40  & 16.80 & 17.33 \\
RA-Agent           & 19.60 & 7.60  & 28.00 & 18.40 \\
Search-o1          & 17.00 & 8.60  & 30.40 & 18.67 \\
R1-base            & 21.00 & 9.80  & 27.78 & 19.53 \\
R1-instruct        & 21.60 & 8.40  & 25.00 & 18.33 \\
Search-R1-base     & 31.20 & 18.20 & 30.56 & 26.65 \\
Search-R1-inst     & 32.80 & 17.40 & 26.39 & 25.53 \\
ZeroSearch-base    & 32.00 & 18.00 & 33.33 & 27.78 \\
ZeroSearch-inst    & 34.60 & {\textbf{18.40}} & 27.78 & 26.93 \\
\midrule
\multicolumn{5}{l}{\textbf{\textit{Our Methods}}} \\
\addlinespace
L0-Scaffold        & 22.03 & 8.33  & 31.20 & 20.52 \\
L0-RL              & {\textbf{40.63}} & 16.60 & {\textbf{57.60}} & {\textbf{38.28}} \\
\bottomrule
\end{tabular}

}

\end{table}%

\begin{table}

\caption{\label{tbl-qa-results-table2}Main results of L0 using different LLMs as backbone.}

\centering{

\centering

\sisetup{table-format=2.2}
\begin{tabular}{l
                S S % HotpotQA
                S S % Musique
                S S % Bamboogle
                S S S % SimpleQA (EM, F1, LJ)
                S S % Avg.
                }
\toprule
\multirow{2}{*}{\textbf{Method}} & \multicolumn{2}{c}{\textbf{HotpotQA}} & \multicolumn{2}{c}{\textbf{Musique}} & \multicolumn{2}{c}{\textbf{Bamboogle}} & \multicolumn{3}{c}{\textbf{SimpleQA}} & \multicolumn{2}{c}{\textbf{Avg.}} \\
\cmidrule(lr){2-3} \cmidrule(lr){4-5} \cmidrule(lr){6-7} \cmidrule(lr){8-10} \cmidrule(lr){11-12}
 & {\textbf{EM}} & {\textbf{F1}} & {\textbf{EM}} & {\textbf{F1}} & {\textbf{EM}} & {\textbf{F1}} & {\textbf{EM}} & {\textbf{F1}} & {\textbf{LJ}} & {\textbf{EM}} & {\textbf{F1}} \\
 \midrule
 \multicolumn{12}{l}{\textbf{\textit{Qwen-3-4B-Thinking}}} \\
\addlinespace
L0-Scaffold & 14.17 & 18.47 & 4.28 & 7.13 & 18.40 & 21.52 & 22.26 & 27.49 & 29.46 & 14.78 & 18.65 \\
L0-RL       & 38.74 & 49.94 & 16.03 & 23.20 & 60.53 & 68.59 & 63.36 & 75.69 & 81.59 & 44.67 & 54.36 \\
\midrule
\multicolumn{12}{l}{\textbf{\textit{Qwen-2.5-7B-Instruct}}} \\
\addlinespace
L0-Scaffold & 22.03 & 29.91 & 8.33 & 14.65 & 31.20 & 36.15 & 21.94 & 28.68 & 30.45 & 20.88 & 27.35 \\
L0-RL       & 40.63 & 52.42 & 16.60 & 25.40 & 57.60 & 68.05 & 61.68 & 75.12 & 80.40 & 44.13 & 55.25 \\
\midrule
\addlinespace
\multicolumn{12}{l}{\textbf{\textit{Qwen-2.5-32B-Instruct}}} \\
\addlinespace
L0-Scaffold & 38.48 & 50.72 & 14.78 & 23.11 & 59.20 & 68.96 & 51.53 & 64.35 & 69.46 & 41.00 & 51.79 \\
L0-RL       & {\textbf{46.14}} & {\textbf{59.16}} & {\textbf{20.22}} & {\textbf{30.03}} & {\textbf{65.07}} & {\textbf{73.70}} & {\textbf{67.58}} & {\textbf{81.28}} & {\textbf{87.29}} & {\textbf{49.75}} & {\textbf{61.04}} \\
\addlinespace
\bottomrule
\end{tabular}

}

\end{table}%

\subsection{Analysis: Impact of Task Difficulty and Dynamic
Sampling}\label{analysis-impact-of-task-difficulty-and-dynamic-sampling}

To better understand the training dynamics of our AgentRL framework and
the challenges posed by complex tasks, we analyze agent behavior under
different data difficulty settings: (1) low-difficulty QA tasks. (2)
high-difficulty QA tasks. (3) high-difficulty QA tasks trained using a
Dynamic Sampling strategy. On the \texttt{easy\_dataset}, the agent
exhibits stable and predictable learning. In sharp contrast, training on
the \texttt{hard\_dataset} without any mitigation strategy leads to
significant training instability and eventual model collapse.

\begin{figure}

\centering{

\pandocbounded{\includegraphics[keepaspectratio]{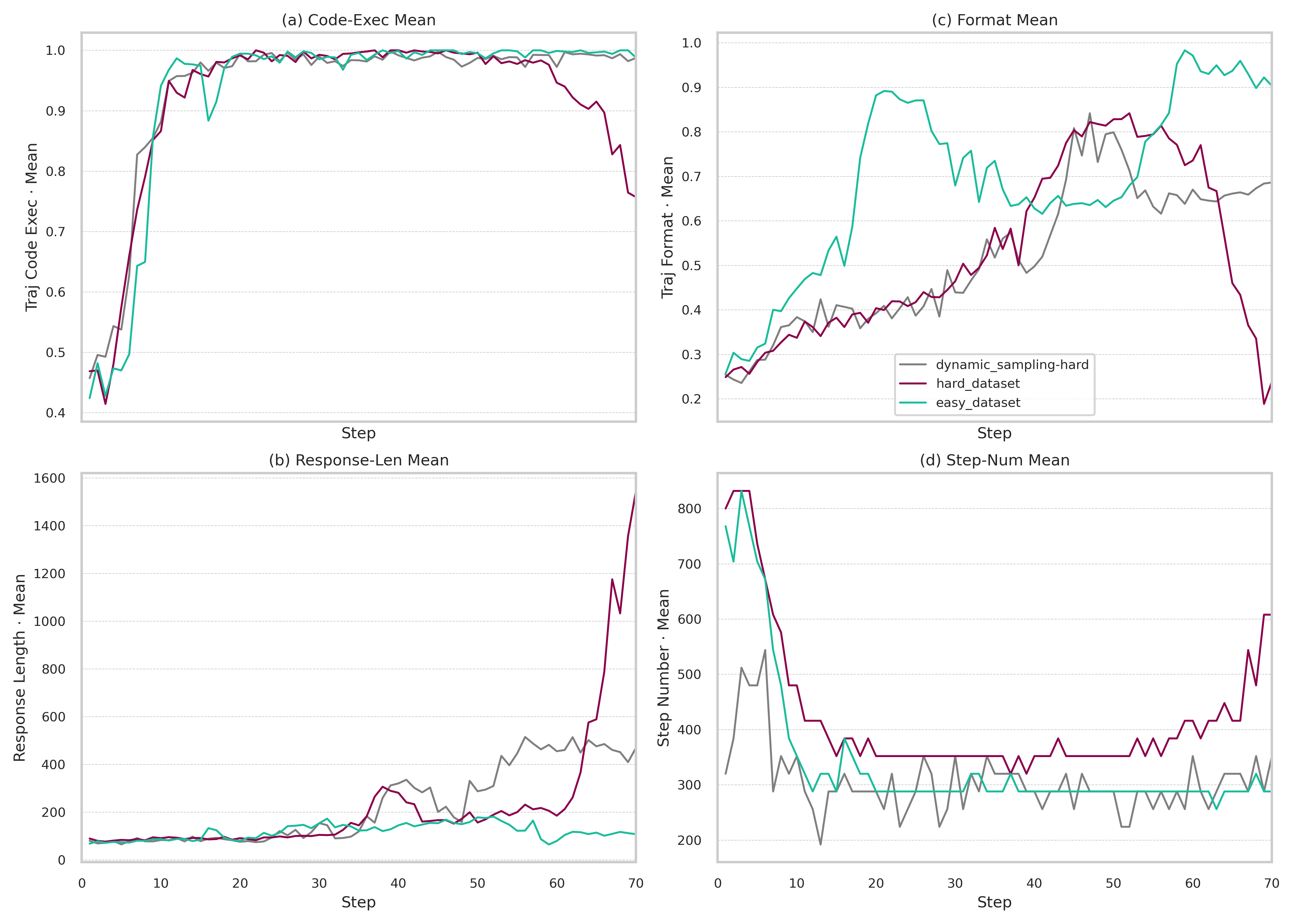}}

}

\caption{\label{fig-abl}The ablation analysis of task difficulty and
sampling strategy.}

\end{figure}

Through RL with a reward on answer and execution accuracy, we observe
that during training the policy quickly learns to eliminate unnecessary
or uninformative queries. For example, the number of redundant search
turns drops sharply as training progresses. It indicates the
self-correction to learn the correct format and begin to eliminate
unnecessary steps, thereby focusing its reasoning and retrieval more
effectively. The average response length in Figure~\ref{fig-abl} (b) and
the number of steps to completion in Figure~\ref{fig-abl} (d) both
explode in the latter half of training. This indicates that the agent
enters a state of inefficient, repetitive looping, generating
excessively long and often incoherent thought processes in a failing
attempt to find a solution. This ultimately leads to a degenerate policy
where the agent can no longer produce well-structured or valid actions,
causing the observed collapse in format and execution rewards. The
\texttt{dynamic\_sampling-hard} configuration demonstrates the
effectiveness of our proposed mitigation strategy. By dynamically
adjusting the sampling of tasks, we can maintain training stability even
on the most challenging dataset. As shown in Figure~\ref{fig-abl} (a)
and Figure~\ref{fig-abl} (c), both the Code Execution and Format
Compliance rewards remain stable throughout the training process,
avoiding the collapse seen in the standard hard\_dataset configuration.

\newpage{}

\section*{References}\label{references}
\addcontentsline{toc}{section}{References}

\phantomsection\label{refs}

\end{document}